\pdfoutput=1

\documentclass[11pt]{article}

\usepackage[preprint]{acl}
\usepackage{booktabs,makecell, multirow, tabularx} 

\usepackage{times}
\usepackage{latexsym}
\usepackage{amsmath}
\usepackage{CJKutf8} 

\usepackage[T1]{fontenc}

\usepackage[utf8]{inputenc}

\usepackage{microtype}

\usepackage{inconsolata}
\usepackage{graphicx}

\def  \be \begin{equation} 
\def  \ee \end{equation} 
\title{GCOF: Self-iterative Text Generation for Copywriting Using Large Language Model}

\author{
Jianghui Zhou$^{*, \dagger}$, Ya Gao$^{*}$, Jie Liu$^{*, \dagger}$, \\
{ \bf Xuemin Zhao, Zhaohua Yang,  Yue Wu,  Lirong Shi} \\
    JingDong Technology \\
    \texttt{\{zhoujianghui, gaoya26, liujie365\}@jd.com}
}

\begin{document}
\maketitle
\begin{abstract}
Large language models(LLM) such as ChatGPT have substantially simplified the generation of marketing copy, 
yet producing content satisfying  domain specific requirements, such as effectively engaging customers,
 remains a significant challenge. 
 In this work, we introduce the Genetic Copy Optimization Framework (GCOF) designed to enhance both efficiency and engagememnt 
 of marketing copy creation. We  conduct explicit feature engineering within the prompts of LLM. 
 Additionally, we modify the crossover operator in Genetic Algorithm (GA), integrating it into the GCOF to enable automatic feature engineering.
  This integration facilitates a self-iterative refinement of the marketing copy.
  Compared to human curated copy,  Online results indicate 
 that  copy produced by our framework achieves an average increase in click-through rate (CTR) of over $50\%$.
\end{abstract}
{\let\thefootnote\relax\footnote{{$*$ Equal contribution}}}
{\let\thefootnote\relax\footnote{{$\dagger$ Corresponding author}}}
\section{Introduction}
\begin{figure}[h]
  \centering
  \includegraphics[width=0.7\columnwidth]{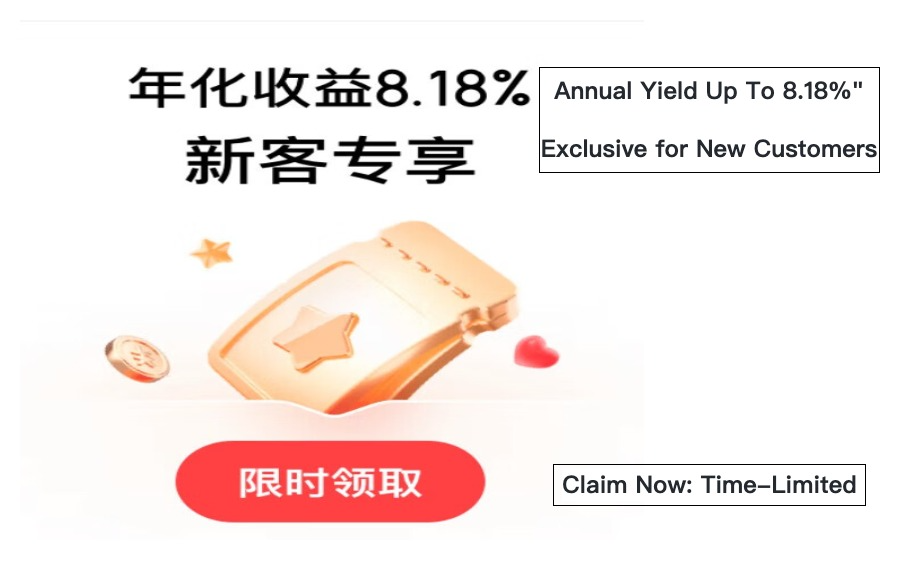}
  \caption{ A marketing campaign card  within JD Finance app.  Promotional copy encompasses three  components: entitlement, target audience and  call-to-action.}
  \label{fig:figure1}
\end{figure}

To drive business growth, operational teams are frequently tasked with initiating marketing campaigns. 
Each campaign necessitates a significant investment of time dedicated to the creation of marketing copy.
 Figure~\ref{fig:figure1} depicts a typical marketing campaign card within the JD Finance app, 
 consisting of  curated promotional copy and  image asset.

Conventionally,  operators rely on user feedback to iteratively refine the copywriting for each campaign. 
This trial-and-error approach is not only inefficient but also exceedingly laborious. Moreover,  there exists 
an opportunity to leverage accumulated knowledge across various campaigns, such as  demographics, preferred writing tokens and tone  of target segment. 
In this study, we introduces the Genetic Copy Optimization Framework (GCOF), an innovative solution for automated generation of marketing copy.
It can generate copies with user specified attributes, and can also make self-refinement giving user feedback iteratively.

Controllable text generation (CTG) enables users to specify  attributes for text production~\cite{ctg_survey}. 
Conventionally, practitioners  resorted to modifying decoding 
strategies~\cite{finetune_attribute, ctrl} or adjusting prompts~\cite{prefix_tuning}.
 They are inflexible in domain transferring and attributes extension.  
With  LLMs, In-Context Learning (ICL) ~\cite{dong2023survey} has improved the generation
 by providing a few illustrative examples. Empirical tests  revealed that the output still often falls short of expectations.   
Chain of Thought (CoT)~\cite{COT}  and  Tree of  thoughts (ToT)~\cite{TOT}  can enhance text generation through iterative interactions.
 However, these approaches incur the cumbersome of multi-round drafting prompts and the increased costs with multiple LLM inferences.

Most of  natural language generation (NLG) models are optimized for next-token prediction, a process misaligned 
with the interests of operational team.  CREATER~\cite{wei2022creater} propose a contrastive 
learning methodology,  aimed at generating high-CTR ad copy. Their framework necessitates the 
aggregation of  substantial data for pre-training and subsequent fine-tuning,  an approach 
impractical for campaigns with limited data. 

GCOF iteratively conducts feature engineering within  prompt  to refine the generation of marketing copy, with 
features of the copy initiated from experiences of operational team, 
 Typical features include  marketing entitlement position and description, copy  
style (urgency,  exclusivity et al.), punctuations  and inductive tokens. 
It has been established that certain tokens and their positions are pivotal in 
the conversion rates of campaign copy. 
Consequently, GCOF is  designed to identify and select keywords, as well as other principal features, which are
essential to conversion rate.  Then GCOF leverage the selected features to generate impressive campaign copy.   

Furthermore,  we introduce  Genetic Algorithm (GA) to find optimal features  for subsequent generation  automatically.  
However,  marketing copy generated with keywords  identified by  GA naively is often suboptimal. The next generation of keywords
are  not controllable, and often lead to copy of  lower CTR compared to that of operator curated. 
We find that the common tokens within the various keyword sets of  marketing copies play a significant role. 
It became essential to retain these commonalities  while exploring the variations within their differences.
 Thus,  we replace the crossover operator in  GA with that derived from 
 differential evolution algorithms (DE)\cite{de1997, Guo2023ConnectingLL}.  
 A reward model whose objective is conversion of the copy is introduced to serve as the  fitness function of GA. 
 To mitigate the data hungry problem, we construct the reward model from LLM.

To summarize, our main contributions include:
\begin{itemize}
\item We present a novel methodology that conducting feature engineering explicitly  in  the prompt of LLM.
Besides demonstrations of  good copy with higher CTR and bad one with lower CTR,  the keywords extracted by LLM are aggregated and selected by LLM to 
enhance  engagement of the  final copy.
we refine the generative process to enhance  engagement of the  copy.
\item Our approach introduces a modified  GA in which the traditional crossover operator is supplanted by one derived from DE. The fitness function is governed by a reward model that leverages GPT-4, with the specific objective of optimizing the CTR of  campaign copy.
\item Online result confirms that our strategy lead  a marked improvement in copy generation. The data illustrates that our automated framework outperforms manually written text, achieving an increase in CTR by $30\%\sim50\%$.
\end{itemize}

\section{Related Work}
\noindent\textbf{Attribute-Based CTG}  is to generate text that is not only coherent  but also aligns with the desired attributes, 
we refer to  ~\cite{ctg_survey} for  survey.  ~\cite{finetune_attribute},  CTRL~\cite{ctrl} and 
StylePTB~\cite{style_ptb}  employ control codes  to generate sentences, while GSum~\cite{gsum} includes keywords and relations to  improve 
 the controllability of PLMs.  For multi-attribute CTG,  PPLM~\cite{pplm} iteratively modifies latent representations of a GPT-2. FUDGE~\cite{fudge} uses an attribute predictor to adjust the output probabilities of a PLM, and  GeDi~\cite{gedi} uses smaller PLMs  to hint a larger one generating sentences that satisfy desirable attributes.  
Prefix-tuning~\cite{prefix_tuning}  keeps PLM parameters fixed while optimizing a small, continuous vector through back-propagation.  
InstructCTG~\cite{instruct_ctg}  incorporates different constraints by conditioning on natural language descriptions and demonstrations of the constraints.  ~\cite{model_arithmetic} propose model arithmetic to express prior CTG techniques as simple formulas and naturally extend them to new and more effective formulations.

\noindent\textbf{CTR prediction models} particularly those for recommendations and advertisements,  have rapidly evolved~\cite{Yang_2022}. 
 However, most researches mainly focus on  user interest
 modeling~\cite{zhou2018deepa, zhou2018deepb, pi2019practice}, domain transfer~\cite{ouyang2020minet, fu2023unified}, 
 interpretability~\cite{li2020interpretable, zhang2022iagcn}. Even with LLM,  approaches typically 
 involve utilizing models like BERT to capture semantic features, which are then feed into 
 neural network to model the similarity between items and users~\cite{lin2023recommender, chen2023survey}. 
Nevertheless, the direct use of user feedback to evaluate the quality of  generated content  remains largely unexplored.
 The work in \cite{Mishra2021TSIAA} merely employs
 a BERT model for a straightforward application in CTR prediction.

\section{Methodology}
In this section, we present the GCOF, a copy generation framework  based on modified genetic algorithm (GA).  To improve the CTR of  copy, we do explicit feature engineering in prompt of LLM.  Following discussions with the operational team, we select  features that are most important, including the placement of marketing entitlement, text style, and keywords. 

For brevity, this article will focus solely on keyword utilization as a case. We define multiple keyword categories in advance. For each copy, we can extract keywords and then categorize them accordingly.  GA is utilized to select  the optimal keywords from previous keyword sets iteratively.
After that, the selected keywords are fed into LLM to generate a potential high conversion rate copy. The overall process can be find in Appendix \ref{sec:copy_gen}.

\subsection{GA in automatic feature engineering}
\begin{figure*}[h]
  \centering
  \includegraphics[width=0.6\paperwidth]{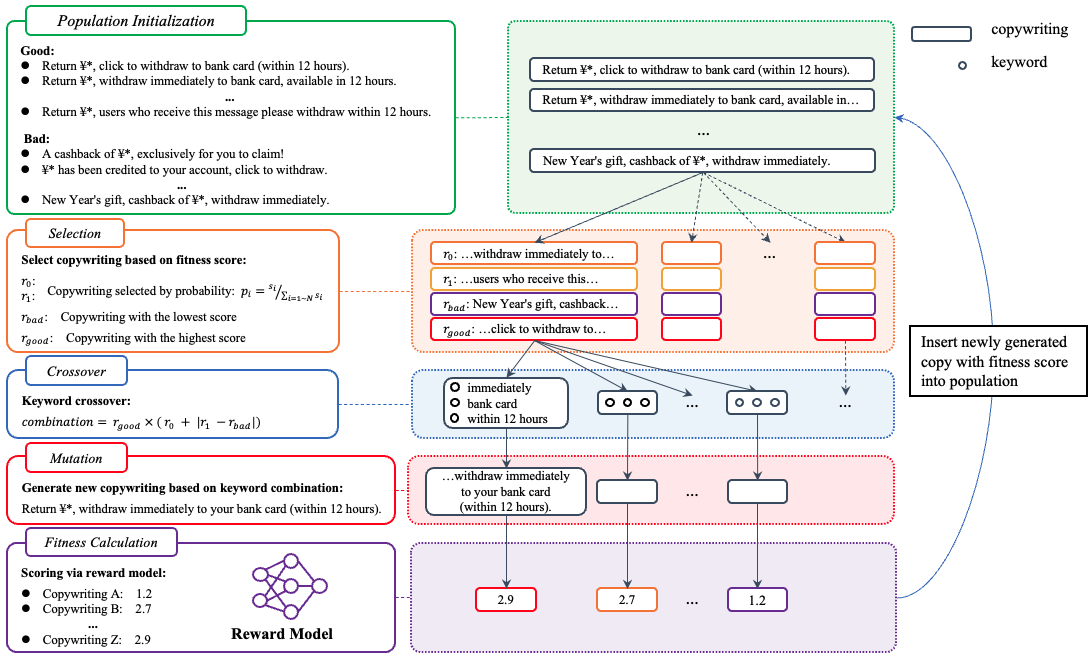}
  \caption{The modified GA pipeline with GCOF Framework: After population initialization, we conduct keyword genetic crossover based on that within DE. The resulting keywords is then fed into LLM to produce the final copy. A reward  model evaluates this output to assign a fitness score.}
  \label{fig:figure2}
\end{figure*}

In our methodology, keywords are conceptualized as genes, and the entirety of  copy is analogous to a chromosome~\ref{fig:figure2}. It is important to note that the gene concept  is extensible and can encompass additional features of marketing copy.

\noindent\textbf{Population Initialization.} The population is initialized with a set of campaign copy candidates from historical  data curated by operators.  We can also employ
LLM to generate copy with keywords extracted from various scenarios.

\noindent\textbf{Selection.} We choose a quartet of copies $\{c_0, c_1, c_{good}, c_{bad}\}$ from the population for each iteration. Copies  $c_0$ and $c_1$  are sampled with weights proportional to their fitness scores: \begin{equation}
p_i = \frac{fs_i}{\sum_{j=1}^N fs_j}
\end{equation}
where  $fs_i$ denotes the fitness score of the $i-th$ copy, that will be elaborated later.  We define  $c_{bad}$  as the copy with the lowest fitness score in the popluation, 
 and conversely,  $c_{good}$   as the one with the highest fitness score.
  
 \noindent\textbf{Crossover.}  For each selected copy, we have one keywords set. Our task is to select keywords  from selected keywords sets as well as fitness score of the corresponding copies.  This crossover operation is informed by that in  DE:
\begin{eqnarray}
 r_{temp} &=& r_0 + |r_1 - r_{bad}|, \\
 r_{final} &=& r_{good} \times r_{temp}. 
 \end{eqnarray}
In  equations, $r_i$ symbolizes the set of keywords contained in copy.  $r_{temp}$ is constructed by subtracting
 keywords exclusive to $r_{bad}$  from $r_0$ and amalgamating keywords uniquely present in $r_1$ into $r_0$ . The resultant  $r_{final}$  is composed of two distinct segments: 
 the first is the intersection $r_{good} \cap r_{temp}$, while the second segment is a random selection from $r_{good} \cup r_{temp} \setminus (r_{good}\cap r_{bad} )$ .

\noindent\textbf{Mutation}. 
In the final step, LLM is instructed to integrate the keywords from $r_{final}$ to generate  campaign copy. An example of  prompt is depicted in second panel of Figure \ref{fig:fig3}.  
The copy is then inserted into population with fitness score.

\subsection{Reward Model}
As conversion rate serves as the critical metric for copywriting, CTR prediction is the reasonable candidate. However, it is usually data-intensive, especially in the 
case of new campaigns where we encounter the challenges associated with cold starts and domain transfer.  
For specified target segment and channel, we argue that deep learning model employed for CTR prediction~\cite{Mishra2021TSIAA} endeavors to comprehend 
semantics and syntax of the training data, including elements like inductive tokens, entitlement positions.

Following LLM-as-a-Judge paradigm \cite{llm_as_judge, selfreward_model},  GPT-4~\cite{gpt4}  is adopted as the foundational model.  We integrate the concept of self-rewarding, guided by the chain-of-thought (CoT) technique, to direct GPT-4 in implementing a ternary scoring system. Please Refer to Appendix \ref{sec:reward_model} for more detail.

To evaluate reward model, , we assess the classification accuracy of  pairs of marketing copy, akin to pairwise learning-to-rank (LTR). 
 For each piece of copy, we obtain its  CTR and  the score predicted by reward model. The prediction is correct for the  pair of copies  if the copy with the higher score  also exhibits a higher CTR. This implies that our scoring model is expected to adhere to the following constraint 
 \begin{equation}
 fs_i > fs_j \Leftrightarrow  ctr_i > ctr_j.
 \end{equation}
Currently, with curated prompt and ICL, the accuracy of reward model is about $0.75$.

\section{Results and Discussions}
The GCOF is now operational online, enhancing copy generation within the JD Finance App across various campaigns. 
The majority of them have experienced an average increase in CTR of $50\%$. This section details two  scenarios of GCOF-applied copy generation.

\begin{CJK}{UTF8}{gbsn}
\begin{table*}[t]
\scriptsize
\centering
\begin{tabular}{c|c|c|c}
\toprule
\textbf{Scenario} & \textbf{Keywords} & \textbf{Generated Copy} & \textbf{CTR} \\
\midrule
\multirow{4}{*}{\makecell[c]{\\SMS}} 
& \{返, 提现\} & \textbf{Human 1}: 返41.73元，去提现 \\& (Cash Back, Withdraw) &(Get 41.73 Cash Back Now! Withdraw) &  $4.90\%$   \\
\cline{2-4}
~& \{送您, 12小时, 提现, 银行卡, 好运\} & \textbf{Human 2}: 送您41.73元，12小时内提现至银行卡，开启好运 \\& (Give, 12 Hours, Withdraw, Card, good luck) &(Give  41.73! Withdraw to Card in 12 Hours and Unlock Good Luck.) &  $4.66\%$   \\

\cline{2-4}
~ & \{返, 12小时, 提现, 银行卡\}& \textbf{Iter 1}:  本单返41.73元，12小时内可提现至银行卡\\ &(Cash Back, 12 Hours, Withdraw, Card)&(Get 41.73 Cash Back on Your Order, Withdrawable to Your Card within 12 Hours) &  $5.32\%$  \\
\cline{2-4}
~ &  \{返, 12小时, 提现,  立即, 银行卡\} & \textbf{Iter 2}: 返41.73元，立即提现到银行卡(12小时内) \\ &(Cash Back, 12 Hours,  Instant Withdraw, Card) & (Get 41.73 Cash Back Now, Direct to Your Card within 12 Hours!) &  $7.18\%$  \\
\midrule
\multirow{5}{*}{\makecell[c]{\\Banner}} 
& \{一键领取, 定投礼包, 专属, 马上领取\} & \textbf{Human 1}: 一键领取定投礼包, 小金库专属福利, 马上领取 \\& (One Click, SIP Gift Pack, Exclusive, Get Now) &(Claim Your  SIP Gift  with One Click! Exclusive For Xiaojinku, Get Now) &  $0.50\%$   \\
\cline{2-4}
~& \{攒钱计划, 定投好礼, 开通\} & \textbf{Human 2}: 开启心愿攒钱计划, 定投享好礼, 立即开通 \\& (Savings Plan, SIP Rewards, Activate) &(Start Your Savings  Plan, Enjoy SIP Rewards. Activate Now) &  $0.20\%$   \\
\cline{2-4}
~ & \{一键领取, 专属, 2.8元, 礼包, 马上领取\}& \textbf{Iter 1}:  一键领取专属福利, 2.8元心愿礼包, 马上领取\\ &(One Click, Exclusive, 2.8 CNY, Gift Pack, Get Now) & (Claim Your Exclusive Benefits with One Click, 2.8 CNY  Gift Pack, Get It Now) &  $0.63\%$  \\
\cline{2-4}
~&  \{点击解锁, 限时好礼, 2.80元 等你拿, 一键领取\} & \textbf{Iter 2}: 点击解锁限时好礼, 2.80元等你拿, 一键领取 \\ &(Unlock, Time-Limited,   2.80 CNY, Claim with One Click) & (Unlock  Time-Limited  Gifts,  $2.80$ CNY Awaits! Claim with One Click) &  $0.74\%$  \\
\cline{2-4}
~&  \{限时好礼, 等你拿, 一键领取\} & \textbf{Iter 3}: 限时好礼, 2.80元等你拿, 一键领取 \\ &(Time-Limited ,  2.80 CNY,  Claim with One Click) & (Time-Limited Gifts,  2.80 CNY Awaits!  Claim with One Click) &  $0.78\%$  \\
\bottomrule
\end{tabular}
\caption{Iterative Optimization of Marketing Copy Using the GCOF Across SMS and Banner Campaign Scenarios. The table presents the initial human-generated copy alongside the iterations produced by GCOF, with their corresponding keywords and observed CTR.}
\label{tab:online-result}
\end{table*}
\end{CJK}

\subsection{User Acquisition Through SMS}
In a user acquisition campaign, JD Finance leverages a strategy whereby a $\textyen 41.73$ cash bonus is disseminated via SMS message, 
incentivizing recipients to install and engage with the JD Finance app.

We have initial two human curated copies. The \textbf{Human 1} copy with the keywords "Cash Back" and "Withdraw" achieved CTR of $4.90\%$, while the second one 
, incorporating a broader set of keywords including "Give," "12 Hours," "Withdraw," "Card," and "Good Luck," resulted in a marginally lower CTR of $4.66\%$.  
GCOF  \textbf{Iter 1} utilized a combination of "Cash Back," "12 Hours," "Withdraw," and "Bank," which led to a higher CTR of $5.32\%$.  \textbf{Iter 2} introduced the term "Instant" alongside the previously successful keywords, resulting in a further improved CTR of $7.18\%$. 

The improvement indicates that the GCOF could discern more effective keyword synergies.  
The inclusion of the keywords "Instant Withdraw" in the final iteration suggests that consumers
 respond favorably to immediate benefits, as evidenced by the substantial increase in CTR. 
This finding is consistent with marketing literature that highlights the effectiveness of urgency and immediacy in promotional content.

\subsection{User Acquisition Through Banner}
In an effort to acquire new customers for the Systematic Investment Plan (SIP) services, the fund department has implemented a promotional initiative. This involves the deployment of a $\textyen 2.8$  SIP gift package, prominently featured on the banner slot of the JD Finance App's homepage (Fig. \ref{fig:figure1}). The primary objective of this campaign is to attract users to activate the SIP service. 

Initial human-generated copy (\textbf{Human 1} and \textbf{Human 2}) yielded baseline CTRs of $0.50\%$ and $0.20\%$, respectively. 
In the first iteration (\textbf{Iter 1}),  GCOF retained high-performing keywords from Human1, such as 'One Click,' 'Exclusive,' and 'Gift Pack,' 
while introducing  '2.8 CNY' into the copy, which increased the CTR to $0.63\%$. \textbf{Iter 2} saw 
the introduction of 'Unlock' and 'Time-Limited'—terms not present in the initial copies—while maintaining the effective 'Claim with One Click' call-to-action,
 further improving the CTR to $0.74\%$. 
 In comparison to the second iteration, \textbf{Iter 3}  produced a more concise and focused copy, focusing on the 'Time-Limited' 
 offer. This strategic refinement yielded a slight increase in click-through rate over the second iteration, achieving the highest  CTR of $0.78\%$.

The GCOF's strategy of combining urgency ('Time-Limited'), exclusivity ('Exclusive'), and direct calls-to-action ('Claim with One Click') with 
a specific price point ('2.80 CNY') proved
 to be particularly effective. 

\section{Conclusion}
In this work, we proposed GCOF to generate copywriting iteratively with customer's feedback.  We introduce GA optimization algorithm into the 
prompt engineering to select keywords from parents and guide the copy generation. 
We also employ LLM as the reward function to select the most promising text that can maximize CTR, the core metric for business. 
In future work we will take a next step to train reward model  of  LLM  across various campaigns and channels which can improve the efficiency and effect of path truncating. 
We will also implement RLHF to explore more state space efficiently. 

\section*{Limitations}
The methodology proposed herein is contingent upon the initial user cohort being informed by the operational a priori knowledge, which encompasses preferences of the target customer 
demographic and established practices for crafting marketing copy. Absent the input from operations, the generation process via the Large Language Model (LLM) may not 
achieve the same rate of iteration. Furthermore, it has been observed within the JD Finance App ecosystem that keywords play a pivotal role in user attraction, suggesting 
a high sensitivity to specific terms. This phenomenon raises concerns regarding the generalizability of our approach to other marketing channels, such as email campaigns or
 push notifications, where the influence of keywords may differ significantly.

\section*{Acknowledgements}
We acknowledge our colleague Youzheng Wu for helpful discussions. We also thank our human annotators and operators.
for their help in labelling and evaluation

\bibliography{llm_ctg}
\appendix
\section*{Appendix}
\section{Copy Generation Prompt}
\label{sec:copy_gen}
\begin{figure*}[h]
  \centering
  \includegraphics[width=0.7\paperwidth]{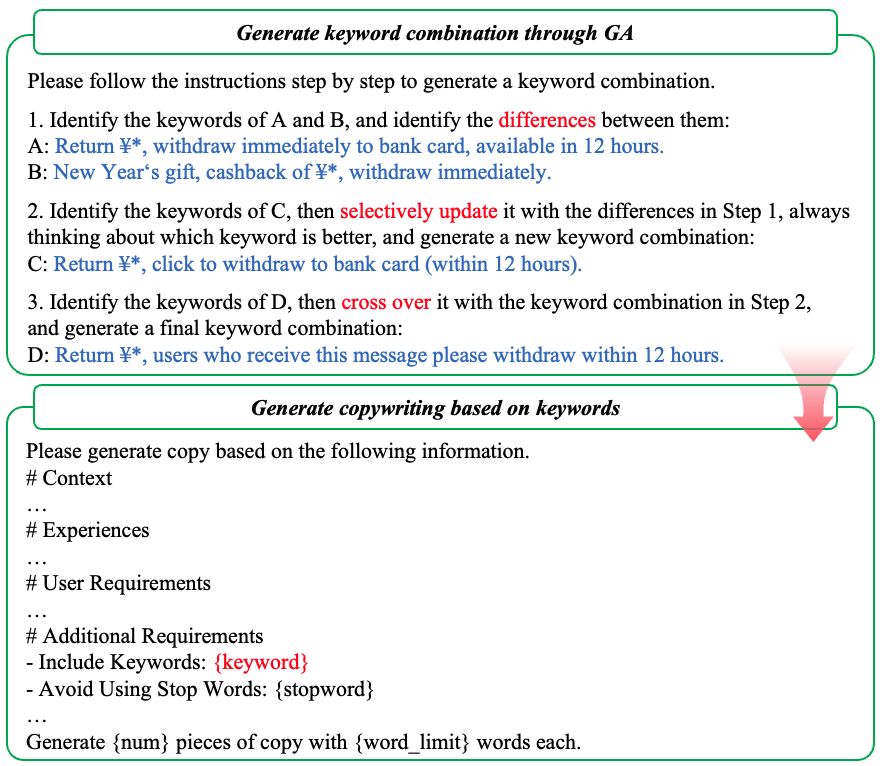}
  \caption{Schematic representation of the two-part prompt process for marketing copy generation using GCOF. The first panel illustrates the keyword combination generation through the Genetic Algorithm (GA), while the second panel depicts the subsequent utilization of these GA-derived keywords within the Prompt Framework to generate the final marketing copy.}
  \label{fig:fig3}
\end{figure*}

In  Fig. \ref{fig:fig3}, we present schematic representation of the two-part prompt process for marketing copy generation using the GCOF. 
The first panel illustrates the keywords selection process with GA. Note that, both selection and crossover are conducted by LLM.
 The second panel depicts the prompt structure of marketing copy generation. Besides the keywords selected from GA, we also tell LLM the domain knowledge and 
  useful experience  from operational team. 

\section{Reward Model}
\label{sec:reward_model}
We leverage LLM to assess the marketing copy for two reasons. Firstly, for the specified channel and segment, what a conventional model learned is the semantic 
and syntax of the sentence. The model makes prediction based on the sentence similarity. Secondly,  from user feedback, we found that certain features, such as 
inductive keywords, punctuations in the sentence,  are crucial in high conversion rate copies. Thus, we  instruct LLM to evaluate copy with three perspectives: 
\begin{itemize}

\item \textbf{Linguistic Expression:} The copy should be of appropriate length, concise, and articulate, with clear and succinct language. Persuasive phrasing is likely to enhance user engagement and marketing effectiveness.

\item \textbf{Logical Structure:}  The copy should exhibit reasonable syntax, with accurate logical relationships such as cause and effect and chronological order. The positioning and organization of marketing elements such as entitlements, timing and calls to action should be rational and precise.

\item \textbf{Information Density:}  The copy should be content-rich, organizing and pairing as many keywords as possible in a sensible manner to provide a high volume of information, thereby delivering a sense of impact to the user.
\end{itemize}

LLM  is required to independently and objectively evaluate the given marketing copy, considering the aforementioned criteria. Specifically, for each criterion,  LLM should make 
detailed thought before  a score ranging from 0.0 to 1.0 being output. Furthermore, we have also give some examples and corresponding explanation to guide the evaluation.
 In this way, we expect LLM can conducting a fine-grained analysis and inference on the copy.

\begin{figure*}[h]
  \centering
  \includegraphics[width=0.65\paperwidth]{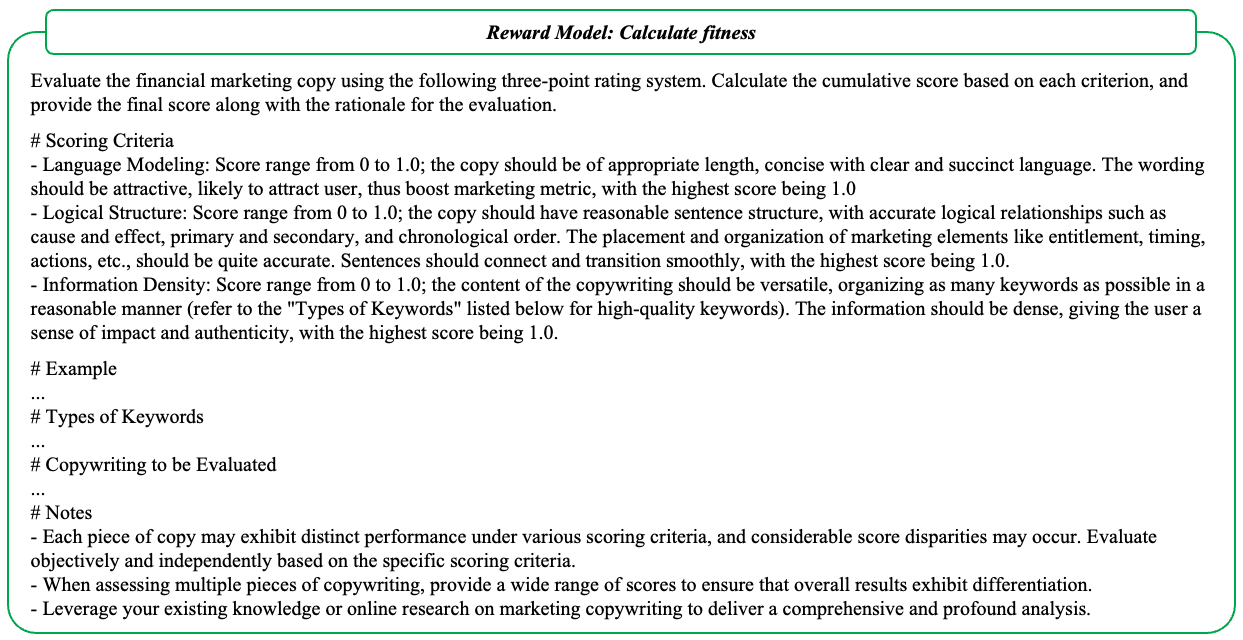}
  \caption{Depiction of the prompt submitted to the Large Language Model (LLM) for generating a reward model, detailing the scoring criteria applied to evaluate a piece of marketing copy.}
  \label{fig:reward_model_prompt}
\end{figure*}

\end{document}